\documentclass[journal,twoside,web]{ieeecolor}
\usepackage{tmi}
\usepackage{cite}
\usepackage{amsmath,amssymb,amsfonts}
\usepackage{graphicx}
\usepackage{subcaption}
\usepackage{caption}
\usepackage{textcomp}

\newcommand{\transformershort}{TransCeption }
\newcommand{\githublink}{\href{https://github.com/mindflow-institue/TransCeption}{GitHub}}

\usepackage{hyperref}
\hypersetup{
	colorlinks=true,
	linkcolor=blue,
	filecolor=cyan,
	urlcolor=magenta,
}
\usepackage[nameinlink]{cleveref}
\usepackage{orcidlink}

\usepackage{algorithm}
\usepackage{algpseudocode}

\def\BibTeX{{\rm B\kern-.05em{\sc i\kern-.025em b}\kern-.08em
    T\kern-.1667em\lower.7ex\hbox{E}\kern-.125emX}}
\begin{document}
\title{Enhancing Medical Image Segmentation with TransCeption: A Multi-Scale Feature Fusion Approach
}
\author{%
Reza~Azad\,\textsuperscript{\orcidlink{0000-0002-4772-2161}}, 
Yiwei~Jia\,\textsuperscript{\orcidlink{0000-0002-5824-8821}},
Ehsan~Khodapanah~Aghdam\,\textsuperscript{\orcidlink{0000-0002-2849-1070}},
Julien~Cohen-Adad\,\textsuperscript{\orcidlink{0000-0003-3662-9532}},
and~Dorit~Merhof\,\textsuperscript{\orcidlink{0000-0002-1672-2185}}%
\thanks{(Corresponding author: D.~Merhof.)}
\thanks{R.~Azad and Y.~Jia are with the Institute of Imaging and Computer Vision, RWTH Aachen University, 52074 Aachen, Germany (e-mail: \{reza.azad; yiwei.jia\}@lfb.rwth-aachen.de).}
\thanks{E.~Khodapanah~Aghdam is with the Department of Electrical Engineering, Shahid Beheshti University, Tehran 1983969411, Iran (e-mail: ehsan.khpaghdam@gmail.com).}
\thanks{J.~Cohen-Adad is with the Mila, Quebec AI Institute and also with the Functional Neuroimaging Unit, CRIUGM, University of Montreal, Canada, and also with the NeuroPoly Lab, Institute of Biomedical Engineering, Polytechnique Montreal, H3T 1J4 Montreal, Canada (e-mail: jcohen@polymtl.ca).}
\thanks{D.~Merhof is with the Institute of Image Analysis and Computer Vision, Faculty of Informatics and Data Science, University of Regensburg, 93053 Regensburg, Germany, and also with the Fraunhofer Institute for Digital Medicine MEVIS, 28359 Bremen, Germany (e-mail: dorit.merhof@ur.de).}
\thanks{This work has been submitted to the \textbf{IEEE} for possible publication. Copyright may be transferred without notice, after which this version may no longer be accessible.}}

\maketitle

\begin{abstract}
While CNN-based methods have been the cornerstone of medical image segmentation due to their promising performance and robustness, they suffer from limitations in capturing long-range dependencies. Transformer-based approaches are currently prevailing since they enlarge the reception field to model global contextual correlation. To further extract rich representations, some extensions of the U-Net employ multi-scale feature extraction and fusion modules and obtain improved performance. Inspired by this idea, we propose TransCeption for medical image segmentation, a pure transformer-based U-shape network featured by incorporating the inception-like module into the encoder and adopting a contextual bridge for better feature fusion. The design proposed in this work is based on three core principles: (1) The patch merging module in the encoder is redesigned with ResInception Patch Merging (RIPM). Multi-branch transformer (MB transformer) adopts the same number of branches as the outputs of RIPM. Combining the two modules enables the model to capture a multi-scale representation within a single stage. (2) We construct an Intra-stage Feature Fusion (IFF) module following the MB transformer to enhance the aggregation of feature maps from all the branches and particularly focus on the interaction between the different channels of all the scales. (3) In contrast to a bridge that only contains token-wise self-attention, we propose a Dual Transformer Bridge that also includes channel-wise self-attention to exploit correlations between scales at different stages from a dual perspective. Extensive experiments on multi-organ and skin lesion segmentation tasks present the superior performance of TransCeption compared to previous work. The code is publicly available on \githublink.
\end{abstract}

\section{Introduction}
\label{sec:introduction}
\IEEEPARstart{B}{enefiting} from the development of deep learning and computer vision technologies, medical image segmentation enables the extraction of meaningful semantic information from raw medical image datasets. It allows precise pixel-wise delineation of anatomical structures and organs with specific shapes, distinct appearances, or varying levels of lesions. Medical image segmentation, one of the key tasks in medical image analysis, is widely used to support clinical applications and computer-aided diagnosis~\cite{azad2022medical,gupta2023segpc,kazerouni2022diffusion}. Several Convolutional Neural Networks (CNNs)-based approaches have been applied to image classification and segmentation with great effectiveness.

The Fully Convolutional Network (FCN)~\cite{long2015fully} enables end-to-end pixel-level semantic segmentation by applying convolutional, activation, and pooling layers in the encoder path and adopting convolutional layers and upsampling in the decoder~\cite{azad2022medical}. The milestone U-shaped encoder-decoder network structure, U-Net~\cite{ronneberger2015unet}, outperformed the state-of-the-art results of numerous medical semantic segmentation tasks~\cite{bakas2018identifying,simpson2019large,heller2019kits19}, and has become the dominant structure for implementing medical image segmentation tasks~\cite{wang2021transbts,azad2021deep}. The encoder in the U-Net structure models the high-level contextual information, and the decoder upsamples the compressed feature to the input resolution. The outputs of each stage of the encoder are connected to the corresponding module of the decoder via skip connections, which helps the decoder to better recover the prediction results by supplementing the spatial information lost in the downsampling process.
The extensibility and symmetry of the U-Net framework allow for a variety of possible designs \cite{azad2019bi}. Compared to the original U-Net structure that simply adds skip connections between stages with the same level of information, UNet3+~\cite{huang2020unet}, and Attention U-Net~\cite{oktay2018attention} allow the decoder to fuse levels of richer semantic information by increasing the number of skip connections or by aggregating feature maps in the skip connections. Differently V-Net~\cite{milletari2016v} network leverages different types of backbones to the U-Net architecture to further facilitate the effectiveness of image segmentation.

Although CNN-based approaches have the promising capability of local feature representation, it is inherently difficult to explicitly model global contextual relationships due to the limited receptive field of the convolutional kernel. This inability to capture long-range dependencies results in suboptimal model performance for complex and intensive prediction tasks. Some works employ stacked convolutional layers and dilated convolutions~\cite{yu2015multi,chen2014semantic,afshin2021multi,azad2020attention} to expand the receptive field or introduce self-attention mechanisms into high-level semantic feature maps. However, this could lead to high model complexity and high computational cost. There is still a need to improve the global interaction capture capability further.

To address the limitation of CNN models in global representation, Vision Transformer (ViT)~\cite{dosovitskiy2020image,azad2023advances} model is proposed to model the long-range dependency using the multi-head self-attention (MSA) mechanism. This method first decomposes the image into a sequence of tokens and then injects the positional embedding into the token sequence as it is fed into the transformer block. It achieves state-of-the-art (SOTA) performance compared to other convolution-based methods.
TransUNet, proposed by Chen et al.~\cite{chen2021transunet}, is the first model that combines Transformer and U-Net for medical image segmentation tasks, where transformer blocks are adapted in the encoder to encode the global information. Some subsequent approaches, such as TransBTS~\cite{wang2021transbts}, CoTr~\cite{xie2021cotr}, and SegTran~\cite{li2021segtran}, also use the CNN-based network as the backbone and integrate the transformer module in some part (e.g., encoder, bottleneck, decoder, or skip connection) to complement the long-range dependencies. Such approaches can be classified as hybrid-transformer architectures. However, these approaches cannot fully combine high-level features and low-level features by using the transformer, and it is still possible to exploit feature correlation between more hierarchical levels. Another kind of architecture can be classified as the pure transformer, such as Swin-Unet~\cite{cao2021swin}, nnFormer~\cite{zhou2021nnformer}, MISSFormer~\cite{huang2021missformer}, DS-TransUNet~\cite{lin2022ds}, TransDeepLab~\cite{azad2022transdeeplab}, and DAE-Former~\cite{azad2022dae}. In particular, the transformer blocks are applied to both the encoder and decoder. Thus, the depth of the model from the input to the output is extended to capture the global feature representations at more levels, with the feature fusion performed in the decoder by skip connections \cite{aghdam2022attention}.

There are some limitations to the previous work: (1) Although the transformer blocks in the hierarchical encoder capture global information at different stages, the transformer only processes the feature layer after patch merging, which uses a single receptive field size. Thus, the multi-scale representation is not properly exploited within each stage. (2) Some contextual bridges with the standard transformer blocks model only the dependencies between token pairs, ignoring the inter-channel interactions. 
Considering the above-mentioned research gaps, we propose the TransCeption model, which aggregates the Inception module into each stage in the encoder and employs a Dual Transformer Bridge to fully exploit the multi-scale feature representations.
Our contributions are as follows:
\begin{itemize}
    \item We propose a multi-scale encoder-decoder architecture TransCeption based on efficient variants of transformer blocks. The multi-scale features of both global contextual information and local finer details are extracted by the novel ResInception Patch Merging (\textbf{RIPM}) and Multi-branch Transformer Block (\textbf{MB transformer}) at each stage of the encoder.
    \item We introduce an Intra-stage Feature Fusion (\textbf{IFF}) module to capture inter-scale dependencies between the output feature maps generated from multiple branches in the MB transformer. IFF emphasizes the interactions across the channel dimension of the concatenated feature maps, while properly preserving the crucial positional information.
    \item We redesign the \textbf{Dual Transformer Bridge} based on the Enhanced Transformer Context Bridge~\cite{huang2021missformer} to further model inter-stage correlations of hierarchical multi-scale features.
\end{itemize}

\section{Method}
\begin{figure*}[ht]
\centering
    \includegraphics[width=\textwidth]{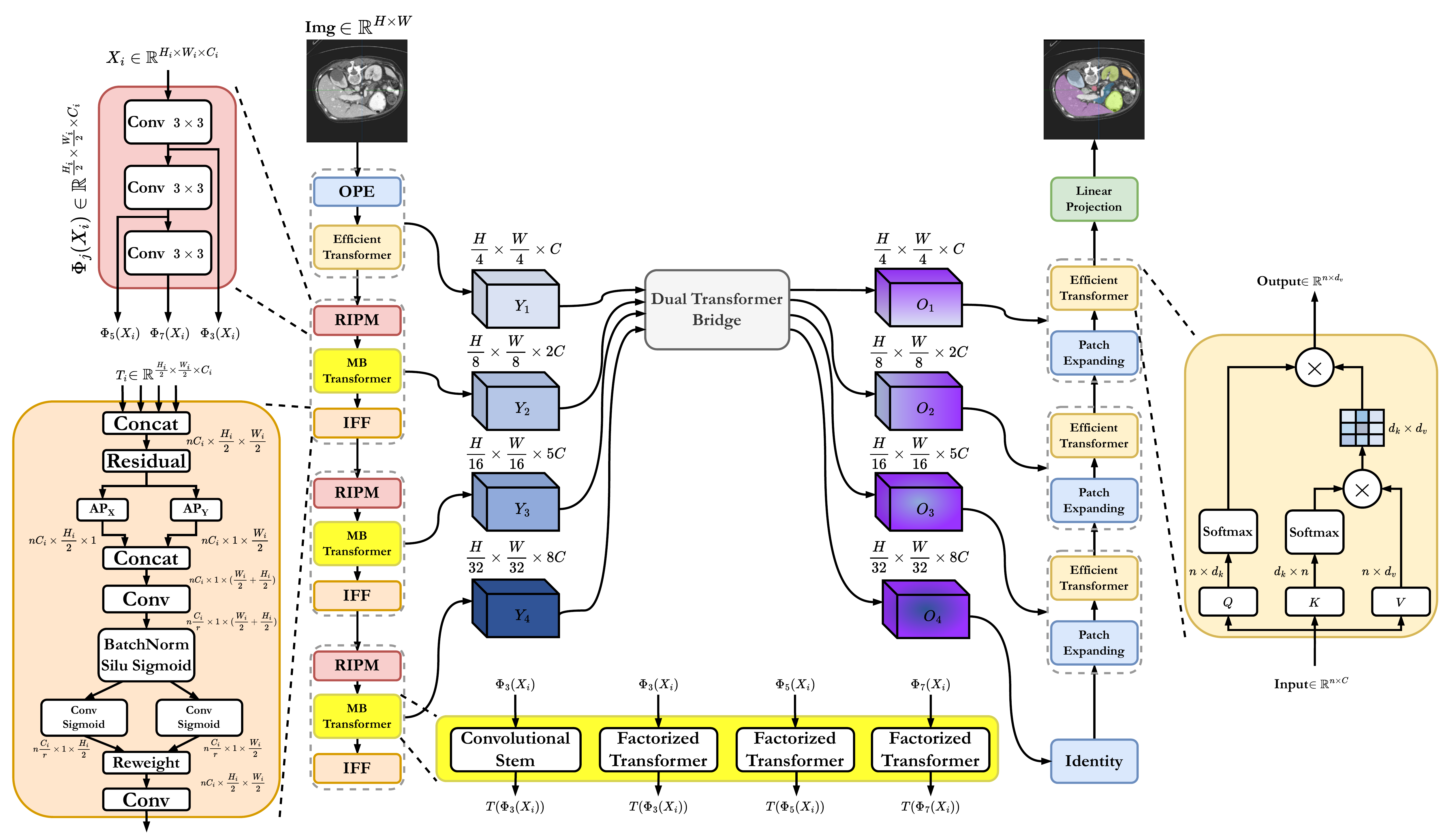}
    \caption{The overview of the proposed TransCeption. The network is composed of the 4-stage encoder, the decoder, and a Dual Transformer Bridge. In the encoder, the first stage utilizes an overlapped Patch Embedding module and an efficient transformer block to capture the low-level semantic features. While the proposed ResInception Patch Merging and MB transformer blocks are aggregated from the second to the deepest stage. The OPE indicates the overlapped patch embedding module.}
    \label{fig:overview}
\end{figure*}

The proposed TransCeption is a U-shaped architecture consisting of the encoder and decoder with a Dual Transformer bridge connecting the two paths. The network overview is shown in \Cref{fig:overview}.
The hierarchical encoder aggregates an overlapped patch embedding module (OPE), three ResInception patch embedding modules, several efficient transformer blocks, and Multi-Branch Transformer blocks followed by the IFF. The decoder includes three efficient transformer blocks and four Patch Expanding blocks to recover the prediction results of the same resolution as the input. The Dual Transformer Bridge learns the inter-scale dependencies from each stage of the encoder and generates the sequence that fuses the hierarchical information from the channel and the token perspectives, respectively. In the next sections, we will describe our method in more detail.

\subsection{Encoder}
\subsubsection{ResInception Patch Merging (RIPM)} \label{sec:RIPM}
In a regular transformer-based U-Net model, the patch merging process is performed at each stage of the decoder using a convolution operation of $P \times P$, and the resulting token sequence is then fed into the transformer block at a later stage. However, patch merging with a fixed patch size at each stage can only capture single-scale features. Therefore, to enhance the representation of the information captured by the patch merging module, we follow the idea of the Inception module and further append two patch embedding branches of $5 \times 5$ and $7 \times 7$ besides the original $3 \times 3$ convolutional patch embedding module. Thus, the embedded feature maps are generated from multiple paths (three in our design) to model the multi-scale representation in each patch merging stage.

Moreover, introducing additional convolutional layers in a parallel manner significantly increases the model complexity, which results in higher computational costs and memory requirements for the hardware. Inspired by~\cite{szegedy2016rethinking,ibtehaz2020multiresunet}, we use the RIPM module to replace the naive Inception module. In particular, two stacked $3 \times 3$ convolutional layers are equivalent to a $5 \times 5$ convolutional layer. Similarly, the three stacked $3 \times 3$ convolutional layers serve as a $7 \times 7$ convolutional layer. This structure can factorize multi-branch inception patch merging with high computational requirements into a smaller and more lightweight model. For example, for the three-branch inception: parallel inception requires $3^2+5^2+7^2 = 83$ parameters, while serial inception requires only $3 \times 3^2 = 27$ parameters.

To constrain the model parameters more effectively, we apply RIPM to the second and subsequent stages of the encoder. Given the input feature map for the $i_{th}$ stage $X_i \in \mathbb{R}^{\frac{H}{2^i}\times\frac{W}{2^i}\times C_i}$ and the projection function with patch size $p$ $\Phi_{p}(\cdot)$, we can obtain the outputs $\Phi_{3}(X_i)$ , $\Phi_{5}(X_i)$ and $\Phi_{7}(X_i)$. The dimensions of the output can be calculated as follows:

\begin{small}
\begin{equation}
    H_i = \left\lfloor \frac{H_{i-1}-p+2d}{s} +1\right \rfloor,\\
    W_i = \left\lfloor \frac{W_{i-1}-p+2d}{s} +1\right \rfloor
\end{equation}
\end{small}

Since the kernel size of the convolutional layer affects the height and width of the output feature map, we adjust the padding size to ensure the consistency of the output resolutions of multiple paths, thus facilitating the subsequent aggregation of multi-scale spatial information in one stage. After normalization and nonlinear activation, the embedding sequences of equal length are passed through the transformer block. This process is repeated three times in the encoder.

\subsubsection{Multi Branch (MB) Transformer Block} \label{sec:transformer_block}
In each MB transformer, the input consists of three feature layers, with kernel sizes of $3 \times 3$, $5 \times 5$, and $7 \times 7$, derived from the ResInception patch merging module. 
To capture the multi-scale representation, we apply the transformer module to each feature map separately. Moreover, to preserve the local semantic representation, we include an additional convolutional kernel that operates on the $3 \times 3$ feature map. Then the output feature sequences from the four sub-modules are concatenated by the IFF module introduced in \Cref{sec:iff}. For the transformer module, we use the CoaT~\cite{Xu2021coat} design.

The main differences between CoaT and the efficient transformer~\cite{shen2021efficient} are the additional use of the convolutional position encoding and the convolutional relative position encoding before and after the attention mechanism, respectively. Besides, CoaT employs a factorized attention mechanism. The difference between factorized attention and efficient attention is that efficient attention computes softmax for both $Q$ and $K$, while factorized attention computes softmax only for $K$:

\begin{equation}
\text{FacAttention} = \rho_k(K)^T V
\end{equation}
\begin{equation}
FA(Q,K,V)= \rho_q(Q)\times \text{FacAttention},
\end{equation}
where $\rho_q$ and $\rho_k$ are normalization functions that can be divided into scaling and softmax functions:
\begin{equation}
\rho_k(K) = Softmax\left(K\right), \quad \rho_q(Q) = \frac{Q}{\sqrt{n}}.
\end{equation}

The original efficient transformer can learn the required global information in a position-free manner with good performance. However, since we need to couple the Transformer module with the ResInception patch merging module, and the multi-path merging produces multi-scale feature maps, each patch invisibly contains local information under different sizes of receptive field mappings and different proximity of connectivity information. Without location encoding, the input token sequence only passes through efficient attention and a feed-forward network, and the output token depends on the corresponding input token by the weights in the computed efﬁcient attention matrix without any awareness of the location feature. We argue that the transformer block involving convolution position encoding can better emphasize local position-based relationships for each branch, supplementing the output token sequence of each branch with positional difference information, and thus enabling richer representation for the result of the multi-branch Inter-stage Feature Fusion module. 
In each convolution position encoding module, the input feature map $X \in \mathbb{R}^{H \times W \times C}$ is first fed into the depthwise convolution with kernel size $3$ and stride size $1$. Then it is flattened to $X_{cpe}\in \mathbb{R}^{N \times C}$ and concatenated with the input residual to serve as the input of the Efficient Attention module.

\subsubsection{Intra-stage Feature Fusion (IFF)} \label{sec:iff}
Inspired by Coordinate Attention~\cite{hou2021coordinate}, we propose our Intra-stage Feature Fusion (IFF) module. The construction of the IFF is shown in \Cref{fig:overview}. Given the $n$ input feature maps $X_{i \in [0, \dots , n-1]} \in \mathbb{R}^{C \times H/8 \times W/8}$ generated from $n$ branches in the MB transformer block, IFF first concatenates them along the channel dimension as the input residual $X_{concat} \in \mathbb{R}^{nC \times H/8 \times W/8}$. Then, global pooling is used to encode the spatial information globally. This approach is in contrast to global average pooling, which collects global information directly from the entire 2D feature map (see \Cref{equ:g_pooling}).

\begin{equation} \label{equ:g_pooling}
    z_g = \frac{1}{H \times W}\sum\limits^{H}_{i=1}\sum\limits^{W}_{j=1}x(i,j).
\end{equation}

The global pooling in the IFF module is factorized into dual-axis feature encoding operations, which can be formulated as follows:

\begin{small}
\begin{equation} \label{equ:h_w_pooling}
    z^h = \frac{1}{W/8}\sum\limits_{0 \le i<W/8}x(h,i), \quad
    z^w = \frac{1}{H/8}\sum\limits_{0 \le j <H/8}x(j, w),
\end{equation}
\end{small}

\noindent where $x$ denotes the input of a single channel, $z_g$ is the output of the global average pooling, and $z^h$ refers to the output of the pooling operation with the kernel $(1, W/8)$ at height $h$. Similarly, $z^w$ is the result of pooling the input $x$ at width $w$ with the kernel $(H/8,1)$.

Global average pooling (\Cref{equ:g_pooling}) squeezes all global spatial information into a value for an individual channel, resulting in the loss of a substantial amount of positional information. Hence, we employ the dual-axis feature encoding (\Cref{equ:h_w_pooling}) to model the feature maps of each channel along both the horizontal coordinate and the vertical coordinate to alleviate the limitations of spatial information preservation. After obtaining the results of the two transformations $z^h$ and $z^w$, which contain a more powerful representation, we aggregate the feature maps by concatenation and a shared $1 \times 1$ convolution layer $Conv_{1 \times 1}$which reduces the channel dimension from $nC$ to $nC/r$, yielding
\begin{equation}
    f = Conv_{1 \times 1}(Concat([z^h, z^w]),
\end{equation}

\noindent where $f \in \mathbb{R}^{C/r \times (H+W)}$ is the squeezed feature map that carries the information features in both horizontal and vertical directions, $r$ denotes the reduction ratio for mitigating the load on the model parameters. Then, $f$ is passed through a normalization layer and a SiLU sigmoid layer to generate $SiLU(BN(f))$. It is split into two tensors $f^h \in \mathbb{R}^{C/r \times H}$ and $f^w \in \mathbb{R}^{C/r \times W}$ along the spatial dimension and transformed through another $1 \times 1$ convolution layer and a sigmoid block to adjust the channel accordingly.

\begin{equation}
    g^h = \sigma(Conv_{1 \times 1}(f^h)), \quad
    g^w = \sigma(Conv_{1 \times 1}(f^w)).
\end{equation}

The processed two tensors $g^h \in \mathbb{R}^{C \times H}$ and $g^w \in \mathbb{R}^{C \times W}$ are utilized as attention weights to re-weight the input residual $X_{concat}$ simultaneously in two directions, yielding the final output of the IFF module.

\begin{equation}
    Y = X_{concat} \times g^h\times g^w
\end{equation}

The proposed IFF module maps the importance between channels of the integrated input feature layer to the attention weights, emphasizing the relationship between multi-scale information at the same stage. It also takes into account the preservation of position information, which reduces the position error in the subsequent encoding process. We argue that applying the MB transformer with this module can facilitate the transfer of rich information more than the original $1 \times 1$ convolutional concatenation, thus boosting the performance of the final segmentation.

\subsection{Dual Transformer Bridge}\label{sec:bridge}
\begin{figure}[!h]
    \centering
    \includegraphics[width=\columnwidth]{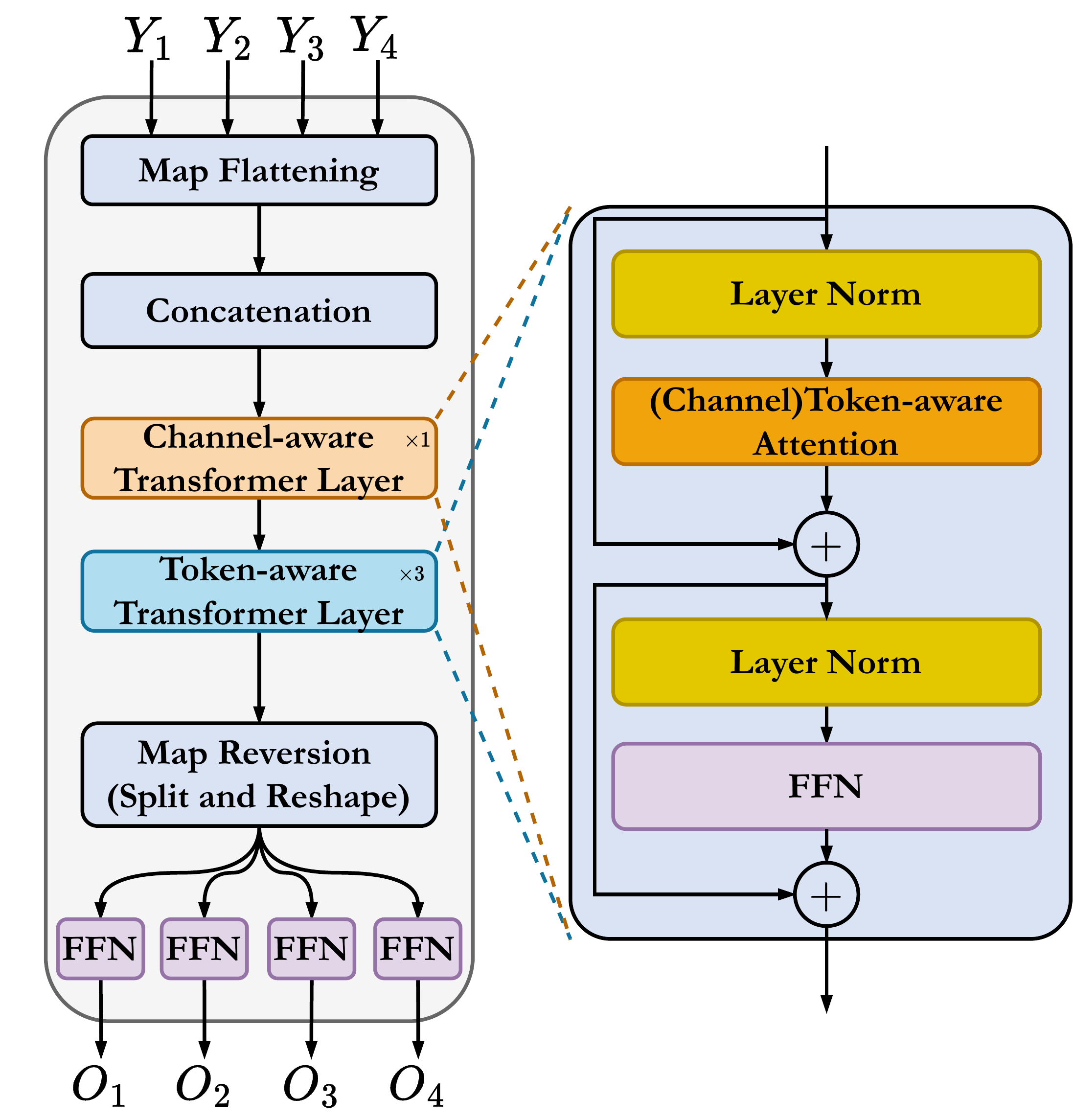}
    \caption{Illustration of Dual Transformer Bridge. The four input feature maps $Y_1$ to $Y_4$ from stages in the encoder are flattened to the token sequences of dimension $N_i \times C$. The token sequences are then concatenated into a long sequence $Y_C$ containing multi-stage information. After channel/token-aware Transformers process the sequence, the output is split and reshaped into four 2D feature maps of the original resolutions.}
    \label{fig:dual_bridge}
\end{figure}

In this section, we will further investigate the multi-scale features with hierarchical relationships between different stages in the encoder. We propose a Dual Transformer Bridge, a bridge structure between the encoder and decoder, in order to extend the scope of multi-scale learning by supplementing the detailed scale-disparity information within a single stage captured by the encoder with large-scale disparity information from multiple stages.
As shown in \Cref{fig:dual_bridge}, after the input patch sequence passes through the efficient transformer block in the first stage of the encoder and the MB transformer blocks in the subsequent three stages, we can obtain a total of four feature map layers, which can be denoted as $Y_i$ where $i$ represents the indices of the stages.

In this paper, we set the number of stages to $4$, so that the Encoder can produce four outputs $Y_1 \in \mathbb{R}^{H/4 \times W/4 \times C}$,$Y_2 \in \mathbb{R}^{H/8 \times W/8 \times 2C}$, $Y_3 \in \mathbb{R}^{H/16 \times W/16 \times 5C}$ and $Y_4 \in \mathbb{R}^{H/32 \times W/32 \times 8C}$. We first perform a reshape operation on these four outputs with different dimensions to expand them into four tensors with the same channel depth, thus facilitating the subsequent concatenation operation along the token dimension. Next, we feed $Y_C$ into the Attention module, which is used to enhance the global dependencies after multi-stage concatenation and to better construct local contextual interactions. In the Dual Transformer Bridge, we propose a sequence-to-sequence Dual Transformer construction to fuse and extract the effective feature expressions contained in the concatenated Token sequence from different perspectives.

\subsubsection{Token-aware Transformer}
For the input $Y_C \in \mathbb{R}^{N \times C}$, we adopt the token-aware attention in the token-aware transformer block, which is equivalent to the original dot product attention module with scale reduction sub-blocks\cite{huang2021missformer} for processing. 
As shown in \Cref{fig:dual_attention}, $Y_C$ is first mapped to $Q \in \mathbb{R}^{N \times d_k}$ . Then we insert the scale reduction module to reduce the shape of the high-resolution feature maps when generating the key $K$ and value $V$. The operation decreases the computational complexity and boosts the efficiency of the attention mechanism. The scale reduction module can be formulated as follows:
\begin{equation}
    K_r = Reshape(K,(\frac{N}{r}, rC))Proj(rC,r).
\end{equation}
$K_{r}$ denotes the new key, which has applied a spatial reduction ratio $r$ to transform the length of pixel sequences from $N$ to $\frac{N}{r}$. Correspondingly, the channel dimension is increased from $C$ to $rC$. Then a linear projection layer restores the channel depth of the intermediate feature layer from $rC$ to $r$. $V_r$ is generated in the same manner:
\begin{equation}
    V_r = Reshape(V,(\frac{N}{r}, rC))Proj(rC,r)
\end{equation}

The dot product attention is synthesized into a tokenized attention matrix by mapping the queries and keys and then allowing each token to be weighted by the coefficients in the attention matrix.
\begin{equation}
    Atten_{T}(Q,K_r,V_r) = Softmax(\frac{QK_r^T}{\sqrt{d_k}}).
\end{equation}

In this way, we can both reduce the computational complexity to $\mathcal{O}(\frac{N^2}{r})$ and learn the inter-dependency between any pair of pixels in the concatenated sequence $Y_C$ by token-wise attention. The representation can be obtained beyond the limitation within a single stage. We can capture the inter-scale relationship information with larger differences between stages.

\subsubsection{Channel-aware Transformer}
In the bridge of our model, we do not only consider a single type of attention mechanism to process the concatenated input $Y_C$. Instead, the context bridge is composed of multiple bridge layers, and each bridge layer contains an attention module. This implies that the input sequence $Y_C$ of large length requires passing through a certain depth (e.g., depth=4 in this paper) of attention blocks when passing through the entire bridge module. This may result in the following limitations: (1) Although token-aware attention performs scale reduction when processing $Y_C$, the spatial complexity and computational complexity are still quadratic to the length of $Y_C$. The model will have a heavy computational overhead if the bridge only employs token-aware attention for all the stages. (2) Token-aware attention only learns contextual information from one perspective of the positional relationship of different scales integrated into $Y_C$. And the inter-channel dependencies are yet to be explored.
\begin{figure}[!th]
	\centering
	\includegraphics[width=\columnwidth]{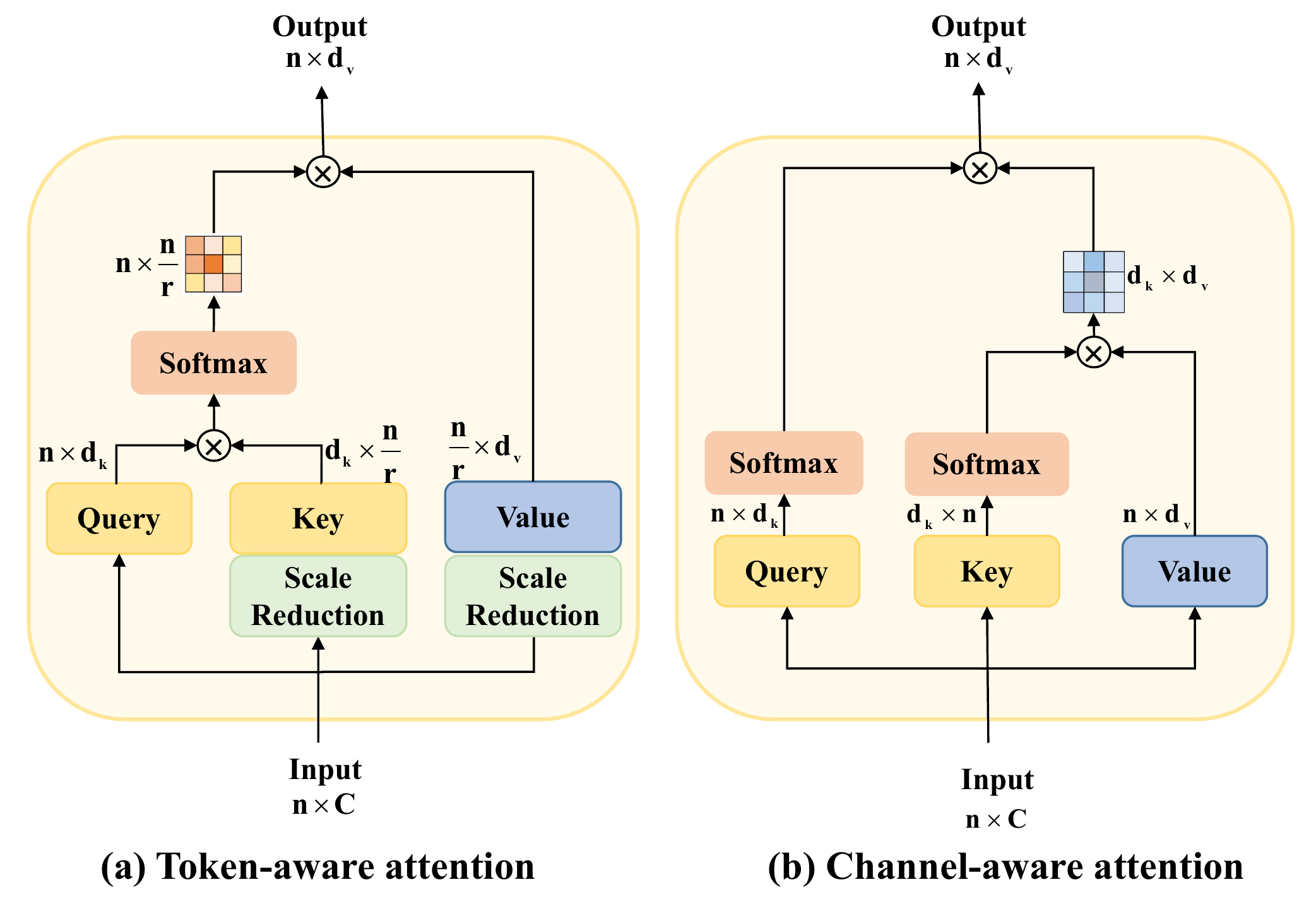}
	\caption{Sketch of two methods of attention mechanism in Dual Transformer Bridge. (a): The token-aware attention is the original dot-product attention mechanism appended with scale reduction sub-modules, where the length of input sequence $n$ is reduced by reduction ratio $r$~\cite{huang2021missformer}. (b): The channel-aware attention mechanism uses a principle similar to efficient attention~\cite{shen2021efficient}, where the $d \times d$ attention matrix is obtained by first computing the inner product of key and value. The whole procedure is mathematically equivalent to the original attention, but with reduced computational complexity.}
    \label{fig:dual_attention}
\end{figure}
Therefore, we adopt the efficient attention mechanism as channel-aware attention, which is illustrated in \Cref{fig:dual_attention}. The input sequence $Y_C$ is first mapped to $Q \in \mathbb{R}^{N \times d_k}$, $K \in \mathbb{R}^{N \times d_k}$ and $V \in \mathbb{R}^{N \times d_v}$. We then aggregate $K^T$ with $V$. Specifically, the vector $k_i^T \in \mathbb{R}^{1 \times N}$ of a single channel is weighted with $v_j \in \mathbb{R}^{N \times 1}$ for all the token elements of the corresponding positions between channels $i$ and $j$, yielding $CA_{(i,j)}$, an element of the channel-aware attention matrix. It summarizes the global semantic information between the two channels. We then weight $Q$ with the $CA$ attention matrix to obtain the final attention output. For the sake of content completeness, the channel attention equation is presented as follows:

\begin{small}
\begin{align}
    &CA = Softmax_{col}\left(\frac{K}{\sqrt{N}}\right)^T \times V\\
    &Atten_{C}(Q,K,V) = Softmax_{row}\left(\frac{Q}{\sqrt{N}}\right) \times CA.
\end{align}    
\end{small}
The spatial complexity of channel attention is $\mathcal{O}(dN+d^2)$ and the computational complexity is $\mathcal{O}(d^2N)$, which are linearly related to the sequence length $N$ of the input $Y_C$. In practice, we have $d_v = d$, $d_k = \frac{d}{2}$ and $d_k\ll N$. This can complement the features extracted from the Token Attention while strengthening the efﬁciency advantage of the bridge module.

The output $Y_{AC} \in \mathbb{R}^{N \times C}$ of the attention module in each bridge sub-layer is normalized and connected to the $Y_C$ residual through the skip connection structure to avoid the vanishing gradient problem, yielding $ResY_{AC}$. We then split the generated sequence and restore both branches to their 2D shapes before the reshaping operation. $Y_i^{sp}$ with their respective heights and widths are passed through the Feed Forward Network (FFN) to generate distinctive representations of the corresponding feature maps. we then reshape the intermediate results into the sequence $M_i$. Finally, $M_i$ is combined with $ResY_{AC}$ via a skip connection and fed back into the FFN module to produce the output of a bridge layer $L$.

\subsubsection{Overview of Dual Transformer Bridge}
All processes can be arranged as follows to illustrate the entire bridge procedure:
\begin{align}
    &Y_i^r = Reshape(Y_i, [B,-1,C])\\
    &Y_C = Concatenate(Y_i^r, dim = 1)\\
    &Y_{AC} = Attention(norm(Y_C)) \label{equ:bri_att}\\
    &ResY_{AC} = Norm(Y_C+Y_{AC})\label{equ:bri_norm}\\
    &Y_i^{sp} = Split(ResY_{AC}, dim = 1)\\
    &M_i = Reshape(FFN(Y_i^{sp}), [B, -1, C])\\
    &L = Concatenate(M_i, dim = 1) + ResY_{AC},\label{equ:end_bridge}\\
    &O_i = FFN(MapReversion(L)),    
\end{align}
where $Attention(\cdot)$ can be set as $Attention_{C}(\cdot)$ or $Attention_{T}(\cdot)$.
Since the Dual Transformer Bridge is stacked by $4$ bridge layers in this paper, we utilize one Channel-aware attention and three Token-aware attention modules subsequently as the optimal configuration. Thus, \Cref{equ:bri_att} to \Cref{equ:end_bridge} perform four times in this process. We also explored other manners of combining the two types of attention blocks, such as interweaving Channel-aware attention and Token-aware attention sequentially or applying the two types of attention blocks in parallel and summing the outputs.

The experiments for the exploration will be demonstrated in \Cref{sec:abla_bridge_config}. After the concatenated sequence from the encoder passes through all the bridge layers, we re-extract the corresponding sequences from $L$ in accordance with the length of each stage of the input, transform them into feature maps $O_i$ of the corresponding resolutions for Stage $i$ in the decoder, and then feed them to the decoder.

\subsection{Decoder}\label{sec:decoder}

The decoder has a symmetric structure corresponding to the encoder consisting of three stages. In contrast to the overlapped patch embedding layer and the ResInception patch merging layer in the encoder, the decoder applies the patch expanding layer to up-sample the encoded features to higher resolution while reducing the channel dimension to half of the input. The block also concatenates the recovered feature map with the skip connection of the same resolution from the encoder. After patch expansion, the feature map is fed into the efficient transformer block to take advantage of the global context information in the decoder.
The resolution of the output is $\frac{H}{4} \times \frac{W}{4}$ after three stages. We finally use a linear projection to restore the resolution as $H \times W$ and a convolution layer to map the dimension to the class number. The output features are the masks of the final segmentation predictions.

\section{Experiments}

\subsection{Dataset}
\subsubsection{Synapse Multi-Organ Segmentation}
The multi-organ Synapse dataset~\cite{landman2015miccai} consists of 30 abdominal CT scans with 3779 axial contrast-enhanced clinical CT images. Each volumetric sample is composed of $85 \sim 198$ slices of the same size of $512 \times 512$ pixels in the entire 3D data. We follow the~\cite{chen2021transunet} setting for the evaluation process. 

\subsubsection{ISIC 2018 Skin Lesion Segmentation}
The ISIC 2018\cite{codella2019skin} dataset, consists of dermoscopic images from diverse populations, which is a large-scale dermoscopic image dataset. The dataset contains 2594 RGB images with a resolution of $700 \times 900$ pixels, and the segmentation map provides the location of the myeloma in each image. The dataset is quite challenging as it contains samples with irregular geometries of the lesion region, varying illumination, different skin textures, and objects such as blisters, dark corners, hair, or rulers as artifacts. 
We follow the same strategy presented in~\cite{alom2019recurrent} to set up the evaluation setting.

\subsection{Implementation Details}
The code of TransCeption is implemented based on Python 3.6. We train the network on an Nvidia RTX 3090 GPU without any pre-trained weights. The size of input images is preprocessed as $224 \times 224$. We randomly apply one to four data augmentations to the input images, including flipping, Gaussian blurring, linear contrast, scaling, rotation, shearing, translation, and Gaussian noise addition. After several comparative experiments, taking into account the GPU memory and the sizes of both datasets, the settings of hyperparameters such as base learning rate (lr), batch size (bs), the maximum training epochs (ep), optimizer (opt), and scheduler (sch) are chosen for each dataset as follows:
\begin{itemize}
    \item \textbf{Synapse}: lr=0.05; bs=16; ep=500; opt=SGD; sch=Cosine Annealing;
    \item \textbf{ISIC 2018}: lr=0.05; bs=24; ep=400; opt=adam; sch=Poly, weight-decay = 1e-4;
\end{itemize}

It should be noted that we follow ~\cite{azad2022transdeeplab} for choosing the optimizer for each dataset. We train our model on the Synapse dataset employing the Cosine Annealing strategy to control the learning rate. The initial learning rate is set to lr and the minimum learning rate to 4e-4. The number of epochs after the restart in the Cosine Annealing scheduler is set equal to the maximum epoch number so that the learning rate decreases monotonically from the beginning of training to the last epoch.


\subsection{Evaluation Results}\label{sec:result_two_dataset}
 We apply the Dice score and 95\% Hausdorff Distance (HD) as our evaluation metric to measure the semantic performance on the Synapse dataset. Dice score, specificity, sensitivity, and accuracy are adopted to evaluate the performance of models on the skin lesion dataset.

\subsubsection{Results of Multi-organ Segmentation}\label{sec:result_synapse}
\begin{figure*}
    \centering
    \includegraphics[width=\textwidth]{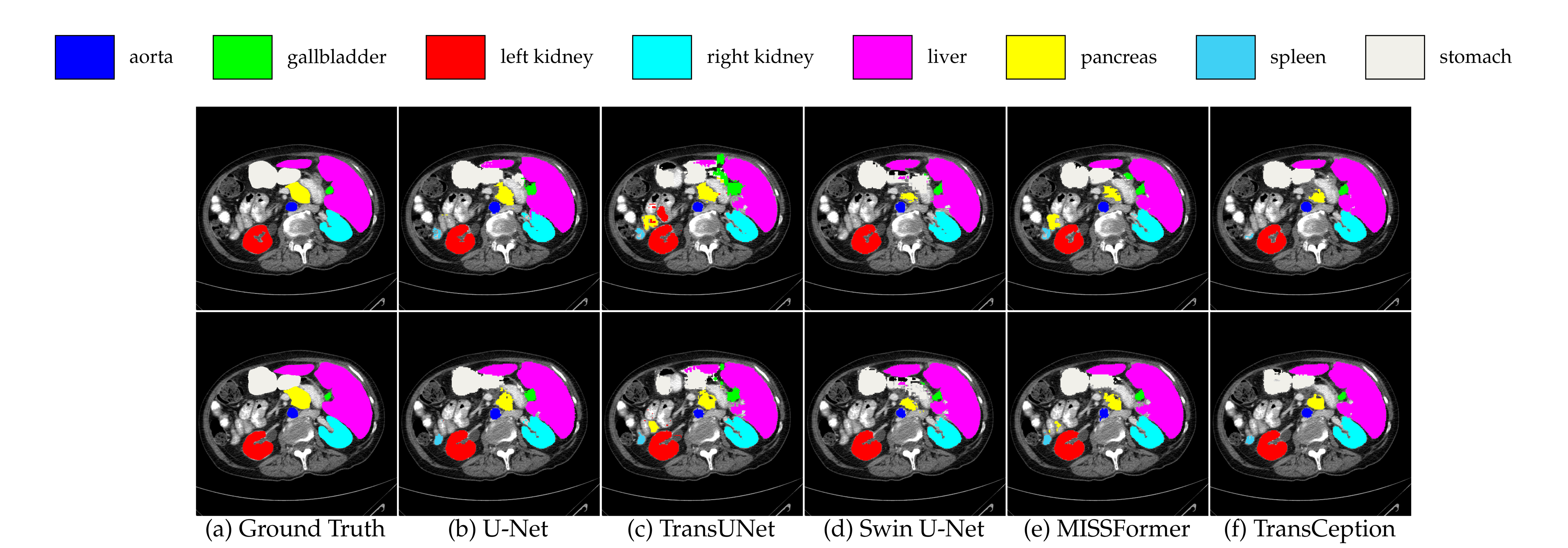}
    \caption{\textbf{Qualitative results} of different models on the \textit{Synapse} dataset. (a) to (f) show Ground Truth, U-Net, TransUNet, Swin U-Net, MISSFormer, and our \transformershort, respectively. In the upper row, \transformershort shows finer boundaries for the region of the stomach and a less false positive prediction mask for the gallbladder compared to Swin U-Net and MISSFormer. In the lower row, \transformershort also shows a smaller false positive region for the pancreas.}
    \label{fig:visual_case_0_99_100}
\end{figure*}

We compare our TransCeption network with six state-of-the-art CNN-based approaches and five transformer-based methods, including V-Net, DARR, U-Net, R50 U-Net, Att-UNet, R50 ViT, TransUNet, Swin-Unet, TransDeepLab, and MISSFormer. TransUNet and Swin-UNet apply the preweight after pre-training on ImageNet. The results of the experiments are listed in~\Cref{tab:er-synapse} with the best results in bold. Our TransCeption achieves the best performance in terms of DSC (82.24\%) and the second-best HD (20.89\%). It is worth noting that TransCeption, as a pure transformer trained from scratch, outperforms all previous pure transformer structures, including Swin-Unet, TransDeepLab, and MISSFormer. This indicates that the contributions of our proposed method for multi-scale fusion at both intra-stage and inter-stage levels lead to a more powerful representation extraction, which is beneficial for improving segmentation performance. The visualization of the results on the Synapse dataset and the explanations are presented in \Cref{fig:visual_case_0_99_100}. 

 \begin{table*}
    \centering
    \caption{Comparison results of some methods on the \textit{Synapse} dataset.}
    \label{tab:er-synapse}
    \resizebox{\textwidth}{!}{
    	\begin{tabular}{l|cc|*{8}c}
    		\hline \textbf{Methods} & \textbf{DSC~$\uparrow$} & \textbf{HD~$\downarrow$} & \textbf{Aorta} & \textbf{Gallbladder} & \textbf{Kidney(L)} & \textbf{Kidney(R)} & \textbf{Liver} & \textbf{Pancreas} & \textbf{Spleen} & \textbf{Stomach} \\
    		\hline 
    		V-Net~\cite{milletari2016v} & 68.81 &- &75.34 &51.87 &77.10 &80.75 &87.84 &40.05 &80.56 &56.98\\
    		DARR~\cite{fu2020domain} & 69.77 &- &74.74 &53.77 &72.31 &73.24 &94.08 &54.18 &89.90 &45.96\\
    		R50 U-Net~\cite{chen2021transunet} & 74.68 & 36.87 & 87.47 & 66.36 & 80.60 & 78.19 & 93.74 & 56.90 & 85.87 & 74.16  \\
    		U-Net~\cite{ronneberger2015unet} & 76.85 & 39.70 & 89.07 & 69.72 & 77.77 & 68.60 & 93.43 & 53.98 & 86.67 & 75.58  \\
    		R50 Att-UNet~\cite{chen2021transunet} & 75.57 & 36.97 & 55.92 & 63.91 & 79.20 & 72.71 & 93.56 & 49.37 & 87.19 & 74.95 \\
    		Att-UNet~\cite{schlemper2019attention} & 77.77 & 36.02 & \textbf{89.55} & 68.88 & 77.98 & 71.11 & 93.57 & 58.04 & 87.30 & 75.75 \\
    	    R50 ViT~\cite{chen2021transunet} & 71.29 & 32.87 & 73.73 & 55.13 & 75.80 & 72.20 & 91.51 & 45.99 & 81.99 & 73.95 \\
    		TransUNet~\cite{chen2021transunet} & 77.48 & 31.69 & 87.23 & 63.13 & 81.87 & 77.02 & 94.08 & 55.86 & 85.08 & 75.62 \\
            TransNorm~\cite{azad2022transnorm} & 78.40 & 30.25 & 86.23 & 65.10 & 82.18 & 78.63 & 94.22 & 55.34 & 89.50 & 76.01 \\
    		Swin-Unet~\cite{cao2021swin} & 79.13 & 21.55 & 85.47 & 66.53 & 83.28 & 79.61 & 94.29 & 56.58 & 90.66 & 76.60\\
    		TransDeepLab~\cite{azad2022transdeeplab}& 80.16 & 21.25 & 86.04 & 69.16 &84.08 & 79.88 & 93.53 & 61.19 & 89.00 & 78.40\\
            HiFormer~\cite{heidari2022hiformer} &  80.39 & \textbf{14.70} & 86.21 & 65.69 & 85.23 & 79.77 & 94.61 & 59.52 & 90.99 & \textbf{81.08} \\
    		MISSFormer~\cite{huang2021missformer} &81.96 & 18.20 & 86.99 & 68.65 & 85.21 & \textbf{82.00} & 94.41 & \textbf{65.67} & \textbf{91.92} & 80.81\\
    		\hline
    		\hline
    		TransCeption & \textbf{82.24} & 20.89 & 87.60 & \textbf{71.82} & \textbf{86.23} & 80.29& \textbf{95.01} & 65.27 & 91.68 & 80.02\\
    		\hline
    	\end{tabular}
    }
\end{table*}
 
\subsubsection{Results of Skin Lesion Segmentation}\label{sec:result_isic}

\begin{figure}[!th]
    \centering
    \includegraphics[width=\columnwidth]{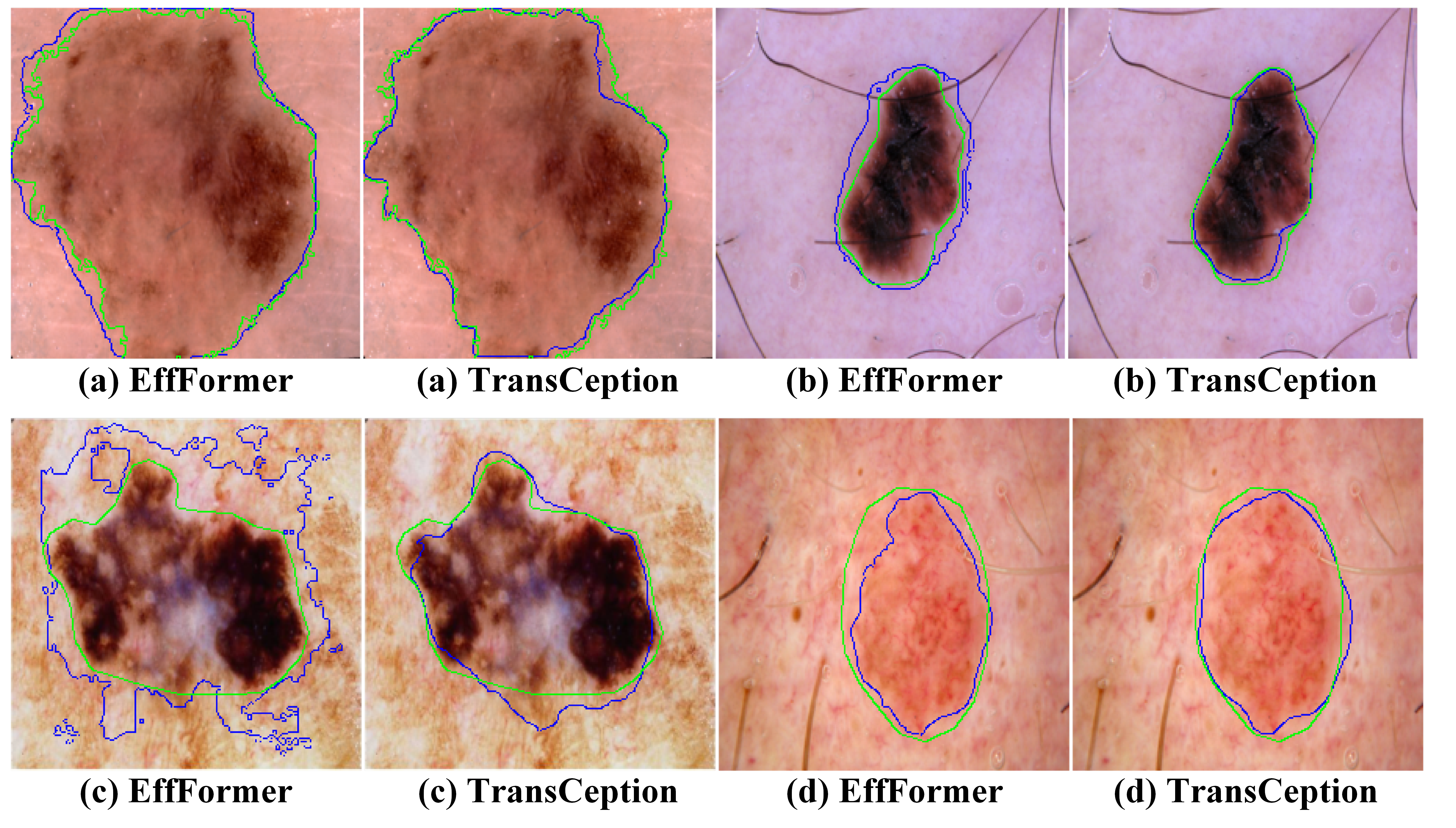}
    \caption[Visual comparisons of EffFormer baseline and our TransCeption model on the \textit{ISIC 2018} Skin Lesion Dataset. Ground truth boundaries are shown in green, and predicted boundaries are shown in blue. TransCeption shows better boundary prediction performance. The contours of masks are closer to that of ground truth with fine-grained details.]{Visual comparisons of EffFormer baseline and our TransCeption model on the \textit{ISIC 2018} Skin Lesion Dataset. Ground truth boundaries are shown in green, and predicted boundaries are shown in blue.}
    \label{fig:isic_visual_case1}
\end{figure}

We evaluate TransCeption on ISIC 2018 skin lesion dataset to demonstrate the generalization of our model and compare the result with other SOTA methods. \Cref{tab:er-isic2018} records the Accuracy (AC), Specificity (SP), Sensitivity (SE), and Dice score (DSC) metrics and the number of parameters of all the methods. Our method achieves the best performance in terms of AC (0.9628), SE (0.9192), and DSC (0.9124). It is worth noting that TransCeption outperforms the pure-transformer network MISSFormer on the ISIC 2018 dataset, which is the closest competitor on the synapse dataset, by 4.67\% of DSC, and also has significant improvements in the AC, SP, and SE evaluation metrics.
\Cref{fig:isic_visual_case1} shows some comparisons of the typical results of the EffFormer baseline and our TransCeption on the ISIC 2018 dataset. From our analysis, it can be seen that TransCeption performs better than EffFormer in the segmentation of skin lesions. The contours of the segmented masks from TransCeption are closer to the ground truth. The predictions are more resistant to noise interference, indicating higher robustness of the model.

 \begin{table}[htbp]
    	\centering
    	\caption{Performance comparison on \textit{ISIC 2018} dataset (best results are bolded). Some results are found in~\cite{azad2022medical}.}
        \resizebox{\columnwidth}{!}{
    	\begin{tabular}{l | c | c c c c  }
    		\hline
    		\textbf{Methods} & 
            \textbf{\#~Params (M)} & 
    		\textbf{DSC} & 
    		\textbf{SE} & 
    		\textbf{SP} & 
    		\textbf{ACC}  \\
    		\hline
    		U-Net~\cite{ronneberger2015unet} & 1.95 & 0.8545 & 0.8800 & 0.9697 & 0.9404 \\
    		Att-UNet~\cite{oktay2018attention} & 34.88 & 0.8566 & 0.8674 & 0.9863 & 0.9376 \\
    		TransUNet~\cite{chen2021transunet} & 105.28 & 0.8499 & 0.8578 & 0.9653 & 0.9452\\  
            TransNorm~\cite{azad2022transnorm}  & 117.48 & 0.8951 & 0.8750 & 0.9790 & 0.9580 \\
            MCGU-Net~\cite{asadi2020multi}  & 27.89 & 0.8950 & 0.8480 & 0.9860 & 0.9550 \\
            FAT-Net~\cite{wu2022fat}  & 28.75 & 0.8903 & 0.9100 & 0.9699 & 0.9578 \\
            TMU-Net~\cite{azad2022contextual} & 165.1 & 0.9059 & 0.9038 & 0.9746 & 0.9603 \\
            Swin-Unet~\cite{cao2021swin}  & 27.17 & 0.8946 & 0.9056 & 0.9798 & 0.9645 \\
            TransDeepLab~\cite{azad2022transdeeplab}& 21.14 & 0.9122 & 0.8756 & \textbf{0.9889} & \textbf{0.9654} \\
    		HiFormer~\cite{heidari2022hiformer} & 25.51 & 0.9102 & 0.9119 & 0.9755 & 0.9621 \\
    		MISSFormer~\cite{huang2021missformer}& 42.46  & 0.8657 & 0.8371& 0.9742 & 0.9453  \\
    		\hline
    		\hline
    		TransCeption  & 38.14 & \textbf{0.9124} & \textbf{0.9192} & 0.9744 & 0.9628 \\
    		\hline
    	\end{tabular}
    	  }
    \label{tab:er-isic2018}
\end{table}

\section{Ablation Study}
\label{sec:guidelines}

\subsection{On the Impact of RIPM and MB Transformer design}
Since the coupling of multiple branches in RIPM and transformer block has numerous possibilities, we conduct several sets of experiments in order to verify the necessity of the coupling method in our structure. To this end, we first evaluate the performance of our baseline model (EffFormer), which only includes patch merging (PM) and a single efficient transformer block in each stage. We further define the TransCeption-RM, which replaces the PM with the RIPM module and uses the multi-branch (MB) transformer instead of a single transformer block. We also investigate the effect of a single transformer block on top of the RIPM module, which we call TransCeption-S. 
In~\Cref{tab:couple_rpe_tr}, different configurations that adopt RIPM and transformer blocks for different numbers of stages are compared. Specifically, "Stages 3" denotes the model with the single-path overlapped patch embedding and the efficient transformer for the first of the four stages in the encoder, and the deepest three stages thereafter using the MB modules. "Stages 4" refers to the model using a combination of RIPM and transformers for all stages in the encoder.
As shown in \Cref{tab:couple_rpe_tr}, we observe that three-stage TransCeption-RM (81.27\%) improves the DSC by 0.60\% over the TransCeption-S (80.67\%). The three-stage multi-path network outperforms the four-stage model in terms of the Dice Score for TransCeption-RM, with a DSC increase of 0.42\%. We argue that the three-stage structure is more advantageous since the four-stage structure has no performance gain, but also makes the parameters of the model more redundant. The results show that the configuration of three-stage RPE and MB transformer models is the most beneficial for segmentation performance.

\begin{table*}[h]
	\caption{Ablation study over \textit{Synapse} dataset to demonstrate the effectiveness and the superiority of various utilized modules.}
	\label{tab:ablation_study}
	\begin{subtable}[h]{0.47\textwidth}
		\centering
		\caption{Effectiveness of RIPM and MB transformers}
		\label{tab:couple_rpe_tr}
		\resizebox{\textwidth}{!}{
		\begin{tabular}[t]{l|*{2}c|*{2}c}
			\hline 
			\textbf{Models} &
			\textbf{Coupling Methods} & 
			\textbf{Stages} & 
			\textbf{DSC~$\uparrow$} & 
			\textbf{HD~$\downarrow$}\\
			\hline 
			EffFormer & PM+single transformer & - & 80.64 & 23.95 \\
			TransCeption-S & RIPM+single transformer & 3 & 80.67 & 21.35 \\
			TransCeption-RM & RIPM+MB transformer & 4 & 80.85 & \textbf{17.81} \\
			TransCeption-RM & RIPM+MB transformer & 3 & \textbf{81.27} & 18.14 \\
			\hline
		\end{tabular}
	}
	\end{subtable}
	\hfill
	\begin{subtable}[h]{0.47\textwidth}
		\centering
		\caption{Configurations of RIPM and MB transformer}
		\label{tab:abl_mb_tr}
		\resizebox{\textwidth}{!}{
		\begin{tabular}[t]{l|*{2}c|*{2}c}
			\hline 
			\textbf{Models} &
			\textbf{Layers} & 
			\textbf{Branches} & 
			\textbf{DSC~$\uparrow$} & 
			\textbf{HD~$\downarrow$}\\
			\hline 
			TransCeption-RM: A & [3,6,3] & [3,3,3] & 80.83 & 20.86 \\
			TransCeption-RM: B & [3,8,3] & [3,3,3] & \textbf{81.20} & 21.07 \\
			TransCeption-RM: C & [3,8,3] & [2,3,3] & 80.52 & \textbf{16.95} \\
			TransCeption-RM: D & [3,6,3] & [2,3,3] & 80.79 & 21.07 \\
			\hline
		\end{tabular}
	}
	\end{subtable}
	\hfill
	\begin{subtable}[h]{0.47\textwidth}
		\centering
		\caption{Impact of IFF module}
		\label{tab:config_iff}
		\resizebox{\textwidth}{!}{
			\begin{tabular}[t]{l|c|*{2}c}
				\hline 
				\textbf{Methods} &
				\textbf{IFF Modules} & 
				\textbf{DSC~$\uparrow$} & \textbf{HD~$\downarrow$}\\
				\hline 
				TransCeption-RM & naive $1 \times 1$ conv & 81.27 & 18.14 \\
				TransCeption-RMI  & 3d concat &  80.15 & 21.14 \\
				TransCeption-RMI & SE & 79.30 & 25.31 \\
				TransCeption-RMI  & SK &  80.40 & \textbf{15.36} \\
				TransCeption-RMI & CBAM & 80.28 & 19.34 \\
				TransCeption-RMI & IFF &  \textbf{81.38} & 20.77 \\
				
				\hline
			\end{tabular}
		}
	\end{subtable}
	\hfill
	\begin{subtable}[h]{0.47\textwidth}
		\centering
		\caption{Various Bridge configuration. c denotes channel-aware attention; t denotes token-aware attention.}
		\label{tab:ablation_bridge}
		\resizebox{\textwidth}{!}{
			\begin{tabular}[t]{l|c|*{2}c}
				\hline 
				\textbf{Methods} &
				\textbf{Bridge Configuration} & 
				\textbf{DSC~$\uparrow$} & \textbf{HD~$\downarrow$}\\
				\hline 
				TransCeption-RMI & w.o. & 81.38 & 20.77 \\
				TransCeption & para &  81.60 & \textbf{16.14} \\
				TransCeption & tttt &  80.67 & 23.98 \\
				TransCeption & ctct &  80.46 & 21.07 \\
				TransCeption & cttt &  \textbf{82.24} & 20.89 \\
				\hline
			\end{tabular}
		}
	\end{subtable}
\end{table*}

\subsection{On the configurations of RIPM and MB transformer}
As discussed above, the cooperation of RIPM and MB transformer provides the model with a performance improvement over the baseline. We performed ablation experiments on RIPM and MB transformers to explore the most optimal parameter configuration.
For the three stages of the encoder using RIPM, we use the branch list to represent the number of branches in each RIPM module. For example, for the branch list $[k_1,k_2,k_3]$, $k_i = 2$ represents the Stage $i$ of RIPM (Stage $(i+1)$ of the encoder) utilizes only $2$ branches of ResInception. One branch of patch merging applies a kernel size of $3$, and the output of the second branch is the result of the stacked two convolutional layers with kernel size $3$ in series. The total kernel size of the stacked convolutional layers is equal to 5. Then the subsequent MB transformer block also integrates two transformer sub-blocks to receive the two outputs from RIPM, and a convolutional sub-block is attached in parallel to process the residual from the kernel size equal to $3$. Similarly, $k_j = 3$ indicates that RIPM constructs the third branch with a total kernel size of $7$ by stacking a third convolutional sub-layer with kernel size $3$ in Stage $(j+1)$ of the encoder, in addition to the two branches mentioned above. Accordingly, the outputs of the three branches serve as inputs to the three transformer sub-modules in the MB transformer module.

In addition to the number of branches, we also investigated the depth of the transformer layers in each transformer sub-module. Layer list $[l_1, l_2, l_3]$ denotes the configuration of the layer number of three MB transformer modules in the encoder. $l_i$ indicates that each transformer sub-block in the MB transformer module is cascaded with $l_i$ sub-layers containing the attention mechanism in Stage $(i+1)$ of the encoder.
It should be noted that the number of layers for the efficient transformer utilized in the first stage of the Encoder is set to a fixed $2$.
The results of the different configurations are listed in \Cref{tab:abl_mb_tr}. The configuration using three branches in all three stages and the transformer layer depths of $3/8/3$ (TransCeption-RM: B) surpasses other combinations of configurations.
The three-path RIPM may be able to enrich the multi-scale receptive field from patch merging, allowing the subsequent transformer to effectively capture the multi-scale long-distance dependencies and obtain informative features.

\subsection{On the impact of IFF module}
Without the IFF module, multi-scale feature maps generated from all the branches in the MB transformer block are simply fused by employing a $1 \times 1$ convolution operation. The outputs are concatenated and projected to the channel dimension for the subsequent requirement. We argue that the capability of the network to model correlations across different scales within each stage can be further enhanced by modifying the concatenation module. 
Therefore, we selected several lightweight attention methods that can emphasize the interactions between channels, and conducted experiments on them (SENet~\cite{hu2018squeeze}, SK Attention~\cite{li2019sknet}, CBAM~\cite{woo2018cbam} and IFF based on CA~\cite{hou2021coordinate}. By adding the IFF module to TransCeption-RM, we can obtain the TransCeption-RMI, which has the architecture without the contextual bridge compared to our proposed TransCeption. 

As shown in \Cref{tab:config_iff}, IFF has the highest DSC among all methods. TransCeption-RMI with integrated IFF module outperforms TransCeption-RM by 0.18\% and 4.22\% with respect to DSC and HD, respectively, which demonstrates the effectiveness of the proposed IFF module. Besides, the performance of other methods decreases rather than increases compared to the naive convolution approach TransCeption-RM, most likely because they ignore the spatial information when extracting features from the depth direction of the channels. As a result, the improvement of inter-channel interactions cannot offset the costs of spatial information. In contrast, the IFF alleviates the lack of positional information by capturing the global information in both the x and y directions. The model integrates the location information and richer feature representation between channels to obtain a more prominent performance than that with the naive convolution concatenation.

\subsection{On various Bridge configurations}
\label{sec:abla_bridge_config}
To confirm the effectiveness of our Dual Transformer Bridge, we compare the TransCeption-RMI architecture employing only skip connection with the TransCeption network appended a contextual bridge for multi-scale fusion. \Cref{tab:ablation_bridge} indicates that our Dual Transformer Bridge (TransCeption $cttt$) improves the network by 0.86\% in the average DSC. 
This result further shows the superiority of the proposed bridge for capturing inter-stage interactions. The bridge allows the segmentation performance to be improved by fusing multi-scale information.

TransCeption exploits a structure that serially concatenates two methods of transformer block, channel-aware transformer (denoted by $c$) and token-aware transformer (denoted by $t$). The channel-aware transformer is embedded in the first of the four bridge layers, and the following three subsequent bridge layers each contain a token-aware transformer. The arrangement is referred to as $cttt$. We explore the impact of several different configurations of our bridge on the performance in order to select the optimal structure for the bridge module. The results of the ablation studies are listed in \Cref{tab:ablation_bridge}.

To demonstrate the enhancement of the dual transformer strategy over the single transformer structure, we evaluate the $tttt$ arrangement where all Bridge layers leverage only token-aware transformers. Compared to the model with the single-transformer Bridge, the insertion of a dual transformer Bridge improves performance from 80.67\% to 82.24\% with regard to DSC. HD also drops from 23.98\% to 20.89\%. It is shown that the dual transformer design is essential for the novel model.

We also investigate the scheme of interweaving the channel-aware transformer and the token-aware transformer ($ctct$). Specifically, the first and third bridge layers employ channel-aware transformer blocks, while the second and fourth layers aggregate token-aware transformers. The results reveal that the $cttt$ scheme outperforms the interweaving $ctct$ strategy by 1.78\% and 0.18\% in terms of DSC and HD, respectively.

We consider whether connecting two approaches of transformers in parallel would be more effective than the fully serial configuration. To this end, we connect the first and second bridge layers in parallel, so that the input of the Bridge model passes through these two layers separately. Then we cascade the two bridge layer output sequences along the dimension of the channel and feed it to the subsequent two serial token-aware transformer layers. \Cref{tab:ablation_bridge} shows that, although the parallel connection approach has a greater gain in model performance than the bridge module with a single transformer (DSC 81.60\% for TransCeption-A, 80.67\% for TransCeption-B), TransCeption-D with the $cttt$ approach obtains the highest DSC (82.24\%) among all bridge configurations, and the corresponding HD is the second best score (20.89\%). Hence, the results confirm that the model adopts the Dual Transformer Bridge with the $cttt$ approach.

\section{Conclusion}
In this paper, we propose a novel architecture for medical image segmentation, namely TransCeption, based on the U-shaped pure transformer framework. It integrates the multi-scale semantic information from intra-stage and inter-stage perspectives to further enhance the good accuracy of prediction. we evaluate our method and conduct ablation studies on the multi-organ dataset and the skin lesion dataset to present the effectiveness of our novel design. We also compare TransCeption with previous work in \Cref{sec:result_two_dataset}. The proposed approach achieves state-of-the-art performance (91.24\% DSC) on the ISIC 2018 dataset and  82.23\% DSC on the Synapse dataset. Visualization is also displayed for qualitative comparison to show the superiority of our method.

\bibliographystyle{IEEEtran}
\bibliography{reference}

\end{document}